\documentclass[letterpaper, 10 pt, conference]{ieeeconf}  % Comment this line out if you need a4paper

\IEEEoverridecommandlockouts                              % This command is only needed if 
\usepackage{graphicx}
\usepackage{color,soul}
\usepackage{hyperref}
\usepackage{mwe}
\usepackage{subcaption}
\usepackage{amsmath}
\usepackage{tabularx} % in the preamble
\usepackage{multirow}

\newcolumntype{L}{>{\centering\arraybackslash}m{3cm}}
\usepackage{float}
\usepackage[thinc]{esdiff}
\usepackage{graphics} % omit 'demo' for real document
\usepackage[table]{xcolor}
\usepackage{caption}
\usepackage{graphicx}
\usepackage{censor}
\usepackage{graphics} % for pdf, bitmapped graphics files
\usepackage{epsfig} % for postscript graphics files
\usepackage{mathptmx} % assumes new font selection scheme installed
\usepackage{times} % assumes new font selection scheme installed
\usepackage{amsmath} % assumes amsmath package installed
\usepackage{amssymb}  % assumes amsmath package installed
\usepackage[noadjust]{cite}
\usepackage{booktabs}
\usepackage{multirow}
\usepackage{algorithmic}
\usepackage{ragged2e}
\usepackage{array}

\newcolumntype{P}[1]{>{\RaggedRight\arraybackslash}p{#1}}
\newcolumntype{M}[1]{>{\centering\arraybackslash}m{#1}}

\usepackage{microtype}
\renewcommand{\baselinestretch}{1.00}

\usepackage[table]{xcolor}

\title{\Large \bf SPROUT: The Open-Source Soft Growing Robot for Search and Rescue}
\author{Antonio Alvarez Valdivia$^{1}$, Ciera McFarland$^{2}$, Robert Reeve$^{1}$, Ankush Dhawan$^{3}$, Chad Council$^{1}$, \\ Megan Richardson$^{1}$, Margaret McGuinness$^{2}$, and Nathaniel Hanson$^{1}$
\thanks{Correspondence: {\tt\footnotesize mmcguinness@nd.edu, nhanson2@mit.edu}}%
\thanks{$^{1}$Lincoln Laboratory, Massachusetts Institute of Technology, Lexington, Massachusetts, USA}
\thanks{$^{2}$University of Notre Dame, Notre Dame, Indiana, USA}
\thanks{$^{3}$Stanford University, Stanford, California, USA}
\thanks{DISTRIBUTION STATEMENT A. Approved for public release. Distribution is unlimited.
This material is based upon work supported by the Department of the Air Force under Air Force Contract No. FA8702-15-D-0001 or FA8702-25-D-B002. Any opinions, findings, conclusions or recommendations expressed in this material are those of the author(s) and do not necessarily reflect the views of the Department of the Air Force.
© 2026 Massachusetts Institute of Technology.
Subject to FAR52.227-11 Patent Rights - Ownership by the contractor (May 2014).
Delivered to the U.S. Government with Unlimited Rights, as defined in DFARS Part 252.227-7013 or 7014 (Feb 2014). Notwithstanding any copyright notice, U.S. Government rights in this work are defined by DFARS 252.227-7013 or DFARS 252.227-7014 as detailed above. Use of this work other than as specifically authorized by the U.S. Government may violate any copyrights that exist in this work.}
}

\begin{document}

\maketitle
\thispagestyle{empty}
\pagestyle{empty}

%%%%%%%%%%%%%%%%%%%%%%%%%%%%%%%%%%%%%%%%%%%%%%%%%%%%%%%%%%%%%%%%%%%%%%%%%%%%%%%%
\begin{abstract}
Soft robots have long been theorized as ideal candidates for use in search and rescue operations; however, there have been significant barriers to entry in graduating these systems from the laboratory to the field. To address this gap in replicable, reliable soft robot systems, we present the designs for SPROUT, the Soft Pathfinding Robotic Observation Unit. The system has matured over years of interaction with professional urban search and rescue communities, with the goal of operating in dusty, wet, and isolated conditions. This manuscript contains supplementary material, including code, parts manifests, CAD assemblies, and build instructions to allow the broader robotics research community to build their own SPROUT systems. The modular hardware and ROS 2-based software stack support task-specific payloads and control functions, allowing SPROUT to be adapted for applications beyond search and rescue, including infrastructure inspection and archaeology. Here, we demonstrate SPROUT performing a variety of challenging inspection and traversal tasks in collapsed structure training sites used by first responders. The main project page is available online at \url{https://sprout-mitll.github.io/sprout/}.

\end{abstract}


%%%%%%%%%%%%%%%%%%%%%%%%%%%%%%%%%%%%%%%%%%%%%%%%%%%%%%%%%%%%%%%%%%%%%%%%%%%%%%%%
\section{Introduction}

In urban search and rescue (USAR) operations, specially trained teams of responders must rapidly assess incidents of structural collapse to triage where trapped victims may be located. Building collapses may lead to the formation of void spaces -- pockets of survivable confined space beneath layers of debris. Current approaches~\cite{murphy2017disaster} that utilize surface-based acoustics or telescoping cameras are inadequate for handling the tortuous paths between a surface entry point and the victim's location.

While rescue robots capable of entering void spaces have been proposed in a variety of forms, spanning tracked ground vehicles \cite{murphy2017disaster} to snake-like robots \cite{Whitman2018uSnake, Ambe2016Kumamoto, Fujikawa2019asc}, these designs struggle in complex, unstructured environments and have hence seen limited deployments.

The subfield of soft robotics yields form factors that have proven useful for confined space navigation. In particular, vine robots \cite{HawkesScienceRobotics2017, CoadRAM2020} are a type of soft robot that navigate via the process of eversion, by which the tip extends and the robot grows in length. 
Eversion and steering \cite{greer2017series} are typically driven by compressed air within a flexible fabric body. Because the body is a pressurized textile rather than a rigid structure, the robot can deform while carrying tools at its tip \cite{jeong2020tip, valdivia2026gotta}. Vine robots have been successfully demonstrated in various field applications, including in-pipe navigation \cite{heap2024large,heap2026transparent,qin20253d}, archaeology \cite{CoadRAM2020}, and even USAR training \cite{der2021roboa, mcfarland2024field}. However, existing robots are either proprietary, dependent on external infrastructure, or too fragile to realistically operate in a disaster-affected region.

\begin{figure}[!tbp]
  \centering
  \includegraphics[width=\linewidth]{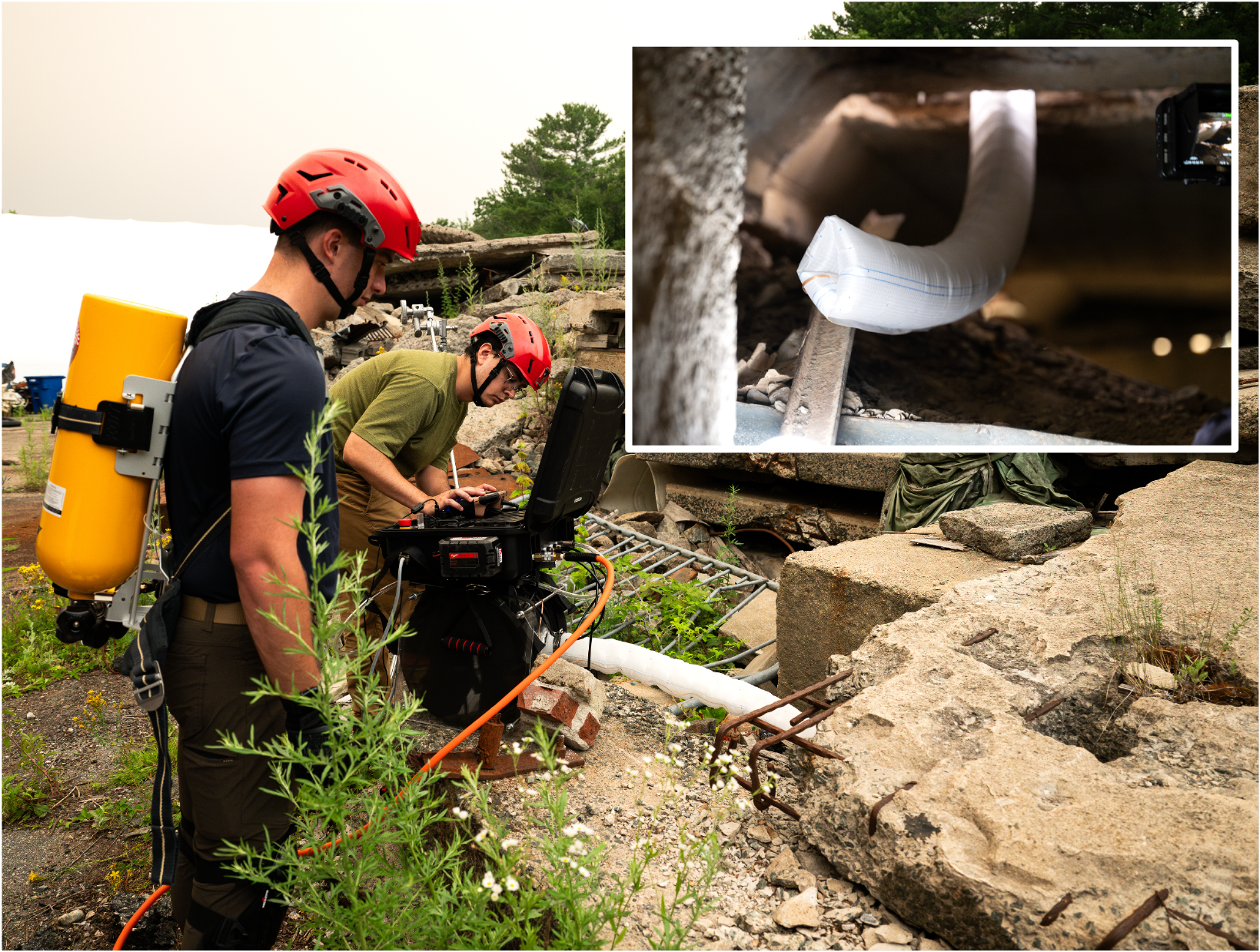}
  \caption{The soft pathfinding robotic observation unit (SPROUT) is an open-hardware, open-software growing vine robot developed to support urban search and rescue (USAR) operations. SPROUT was developed in conjunction with USAR practitioners to meet operational needs while minimizing deployment times. (Inset) Tip of SPROUT vine body exploring the rubble pile.}
    \label{fig:rap_teaser}
    \vspace{-1.00em}
\end{figure}

\begin{figure}[!b]
    \centering
    \vspace{-1.5em}
    \includegraphics[width=\linewidth]{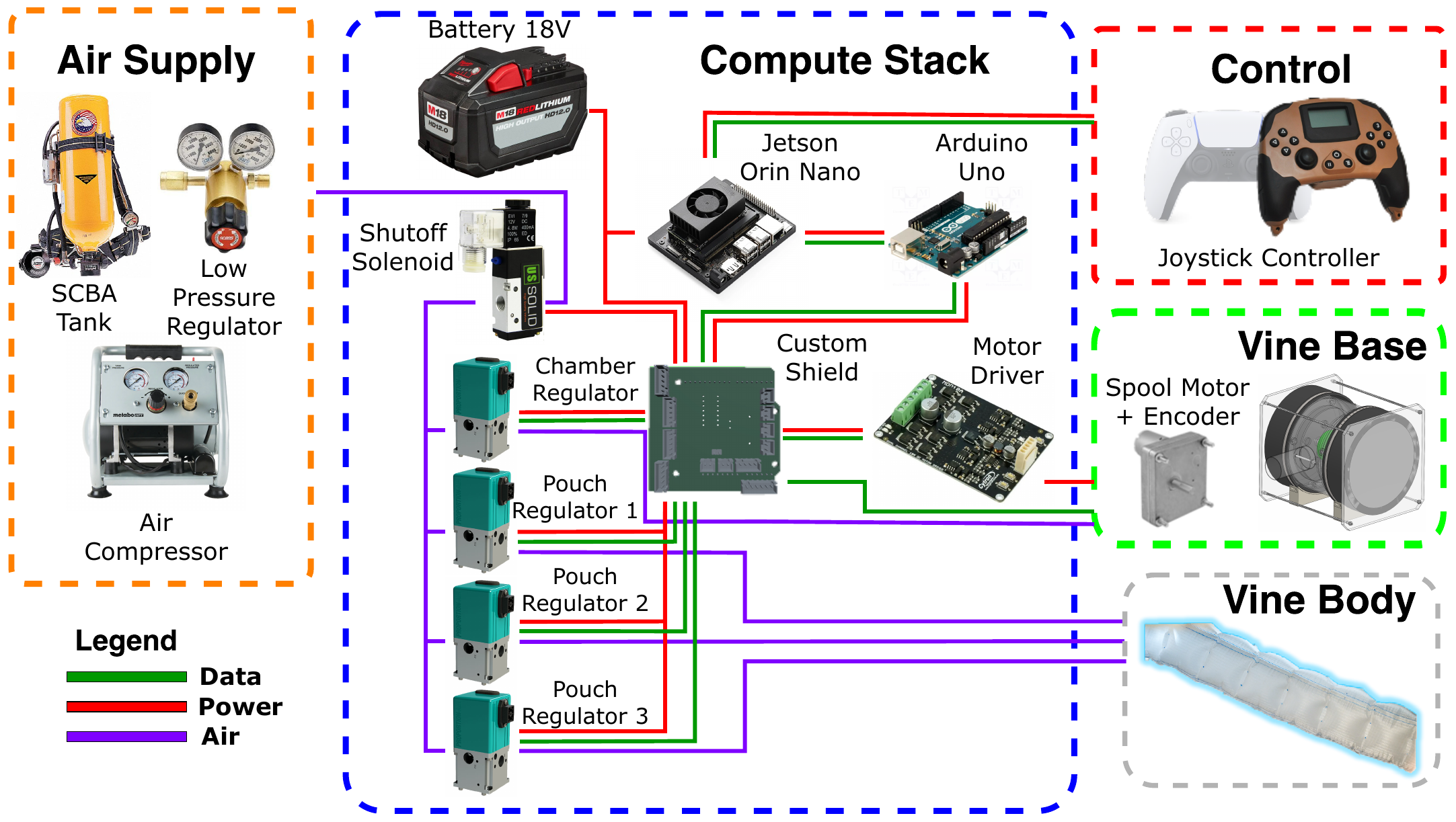}
    \caption{\textbf{System architecture for SPROUT operations}. The robot features an interconnected mixture of pneumatic and electromechanical components to enable portable teleoperation. Pneumatic power can be provided by either an air compressor or a portable air tank. Multiple controller models can also be used to teleoperate the robot.}
    \label{fig:sys_arch}
    % \vspace{-2.00em}
\end{figure}

In order to address these challenges and increase the prevalence of soft robotics in the disaster response community, we present SPROUT: the Soft Pathfinding Robotic Observation Unit. SPROUT has been engineered with USAR teams and validated in austere environments (Fig.~\ref{fig:rap_teaser}). This system encapsulates several years of study \cite{mcfarland2025onsteerability, valdivia2026gotta, mcfarland2024field} and iterative design improvements to yield an easily assembled, portable, encapsulated system ready for field use. This work contributes the following advances to the state-of-the-art:
\newpage
\begin{itemize}
    \item Open-source hardware design for a ruggedized vine base and compute platform construction.
    \item Open-source, modular ROS 2 software stack for teleoperation and data acquisition.
    \item Process for building integrated vine bodies with high pressure ratings and high-curvature steering capability.
\end{itemize}

\section{Design of SPROUT}
SPROUT consists of several interconnected pneumatic, mechanical, electrical, and computational systems to allow the vine robot to grow and steer. The interconnection between these various systems is highlighted in Fig.~\ref{fig:sys_arch}. A full CAD assembly of the system components is shown in Fig.~\ref{fig:cad}.

\subsection{Vine Base}
\begin{figure*}[t!]
    \centering
    \includegraphics[width=1.0\linewidth]{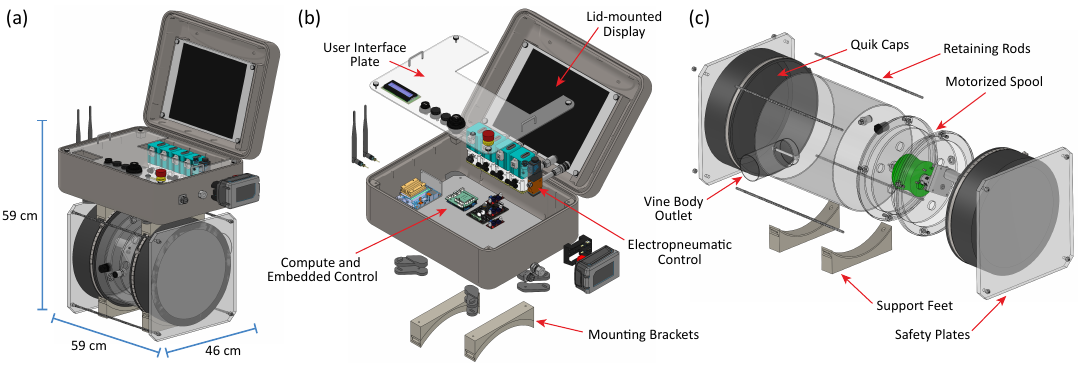}
    \caption{\textbf{SPROUT combines a self-contained compute box with a modified open-source vine base to support portable field deployment.} (a) Assembled SPROUT platform and overall dimensions. (b) Exploded view of the compute box. (c) Exploded view of the vine base. Complete CAD files are provided in the supplementary materials.}
    \vspace{-2.00em}
    \label{fig:cad}
\end{figure*}

The SPROUT vine base (Fig.~\ref{fig:cad}(c)) is derived from the open-source vine robot base described in \cite{CoadRAM2020} and follows the construction procedure available through the \href{https://www.vinerobots.org/}{vinerobots.org} website. 
At a high level, the vine base consists of a cylindrical pressure chamber, a motorized spool, a retraction tendon, an outlet tube, and acrylic side plates that together support the storage, deployment, and retraction of the vine body.
To deploy the robot, the base chamber is pressurized while the motorized spool unwinds the retraction tendon, allowing the vine body to evert through the outlet and grow. Reversing the spool winds the tendon back in to retract the robot. 
The overall geometry and operating principle of the base were retained, while several modifications were introduced to improve assembly, reliability, safety, and portability. The 3D-printed reel core, which forms the center of the motorized spool between the acrylic reel plates, was modified to incorporate heat-set threaded inserts, allowing the plates to be mechanically fastened to the core rather than attached using hot glue. The reel core was also redesigned with a shallow concave groove that passively guides the retraction tendon toward the center of the spool during winding, reducing its tendency to accumulate near the edges. As a safety measure, 12.7~mm thick (1/2~in) acrylic retaining plates are secured to both sides of the base with threaded rods, allowing the system to operate at higher pressures without the risk of separation of the Quik caps. A shoulder strap and carrying handles were also added to facilitate the transport of the assembled system. The supplementary materials include the complete CAD package for the SPROUT vine base, encompassing the full base assembly and all modifications introduced in this work. The original Vine Robots construction instructions \cite{CoadRAM2020} can therefore be used alongside the provided files to reproduce the SPROUT implementation.

\subsection{Compute Box}
To support portable field operations, SPROUT consolidates its computation, power distribution, embedded control, and pneumatic hardware within a portable, weatherproof enclosure that can be transported and installed on the vine base for operation, as shown in Fig.~\ref{fig:cad}(b). The compute box is built around a protective case (930, NANUK) and contains the onboard computer, low-level controller, motor drive electronics, pneumatic pressure regulators, power conversion hardware, and external communication interfaces.
The compute box electronics are mounted to a removable ABS plate secured to the bottom of the enclosure using custom 3D-printed brackets with heat-set inserts. This plate carries the compute and embedded-control hardware, motor driver (MDD10A, Cytron), power-distribution hardware, and electropneumatic components, including the pressure regulators (QB3, ProportionAir) and a solenoid valve. A removable acrylic interface plate covers the internal hardware and supports the user-facing peripherals (USB, Ethernet, HDMI), LCD, power button, and emergency-stop button. A separate display mounted in the enclosure lid connects to the computer via HDMI and provides the primary operator display.

The compute box is powered by an external 18-V battery source (Milwaukee M18), with DC-DC converters providing the required auxiliary and compute voltage rails; the power branches are independently fused to isolate subsystem faults. Complete component specifications, wiring diagrams, connector pinouts, enclosure preparation instructions, and assembly procedures are provided in the supplementary materials.

\begin{figure}[!b]
    \centering
    \vspace{-1em}
    \includegraphics[width=1.0\linewidth]{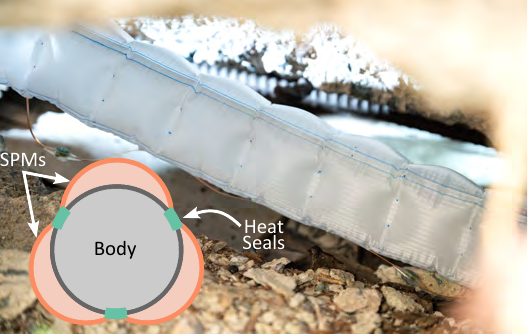}
    \caption{\textbf{Vine Robot Body.} SPROUT uses an integrated-pouch vine body with three longitudinal series pouch motors (SPMs) arranged circumferentially around the body.}
    % \vspace{-2.00em}
    \label{fig:vine}
\end{figure}

\subsection{Software Stack}
SPROUT uses an NVIDIA Jetson Orin Nano for onboard computation and an Arduino Uno Rev3 for low-level actuator and sensor interfacing. The Arduino interfaces with the reel motor driver and the electronic pressure regulators, while the Jetson provides higher-level computation, joystick input processing, and a communication interface. To increase the modularity of the system, SPROUT uses a ROS 2 Humble software stack \cite{macenski2022robot}. The system is configured to launch from power-on, resulting in quick setup and deployment times. The code is structured around a series of nodes that publish standard ROS 2 message types as well as custom messages specific to the vine hardware state.

The Jetson and Arduino devices communicate over a serial connection. The custom Arduino shield exposes additional analog and digital pins which can be used to incorporate additional sensors and low-level hardware onto the robot, such as the joystick demonstrated in \cite{el2018development}.

\subsection{Pneumatic Power}

Previous vine robot designs have largely used standalone air compressors to provide pneumatic actuation. However, in post-disaster environments, air compressors often require 120V AC power while generating noise during operation. Both of these factors are non-ideal for field use. SPROUT is designed to primarily operate on Self Contained Breathing Apparatus (SCBA) air tanks, which are routinely carried by first responders in their equipment loadout, with the option to switch to an air compressor if needed, using quick disconnect hoses. With SPROUT, we use industrial tanks (Scott Industrial SCBA 2216, 3M). A low-pressure regulator (SCBAS-HPR-1-400, SCBAS, Inc.) is used to reduce the tank pressure from 15.3 MPa (2216 PSI) to 207 kPa (30 PSI), as expected by the inlet to the pressure regulator for the base. This setup provides one hour of continuous operation before a tank swap is required. The SCBA tank is backpack mounted and can be carried with the robot, as shown in Fig.~\ref{fig:rap_teaser}.

\subsection{Control Paradigms}

SPROUT is teleoperated using a single joystick that commands steering in a plane perpendicular to the everted body. Joystick motion adjusts the pressure supplied to three series pouch motors (SPMs) \cite{niiyama2015pouch}, soft pneumatic steering actuators arranged around the circumference of the vine body that shorten when pressurized, causing the body to curve toward the actuated side.

The joystick also provides control over the growth and retraction speed, as well as the pressure within the main vine body. We provide two options for joystick integration: a PlayStation 5 controller (Sony) and a Wired Tactical controller (Fort Robotics). The latter is preferred because it is IP66-rated and can withstand the dust and moisture that may be present at a rubble site; however, the PS5 controller is a cost-effective alternative. Since the system uses standard ROS 2 messages, many other control peripherals can also be integrated easily.

\begin{figure*}[!t]
    \centering
    \includegraphics[width=\linewidth]{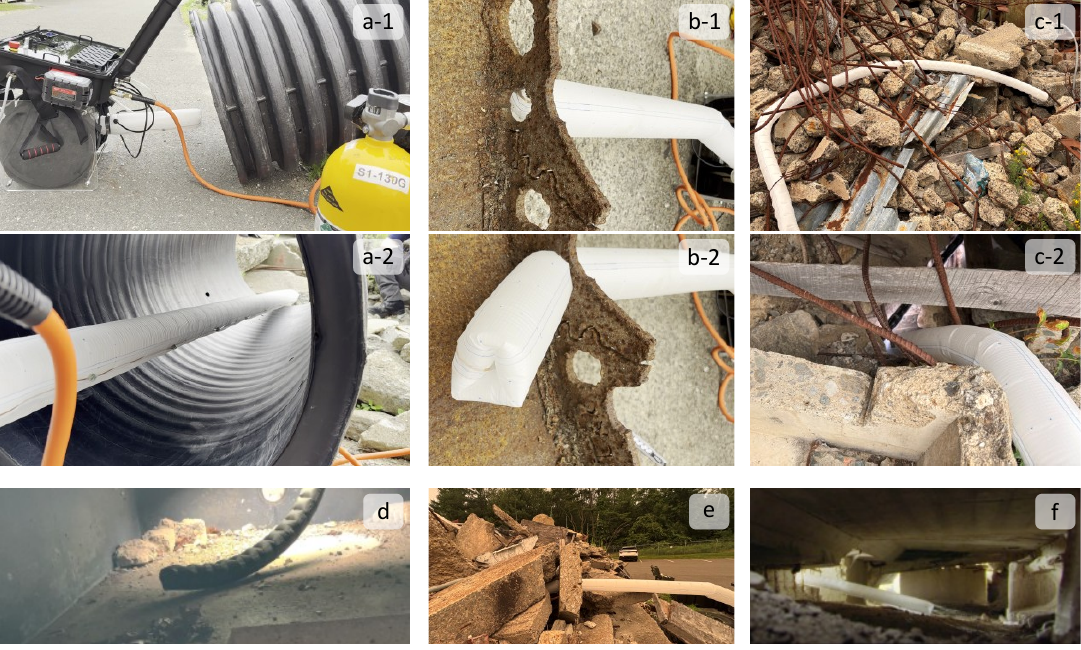}
    \caption{\textbf{Key capabilities of SPROUT in rubble pile.} (a) Vine growing through sloped culvert. (b) Vine growing through rough cut hole in steel plate ($\approx 3.81~\text{cm}$ diameter) smaller than nominal vine diameter 7.62 cm (3 in) and turning after movement through hole (b-2). Subfigures (c-f) demonstrate SPROUT's robustness in navigating the rubble pile including through rebar (c), from vertical entry into void (d), over the surface and through a small entry hole (e), and into a void space (f).}
    \label{fig:experiments}
    \vspace{-2.00em}
\end{figure*}

\subsection{Vine Construction}
The SPROUT configuration presented in this work uses the integrated-pouch vine robot construction described in \cite{mcfarland2025onsteerability}, as shown in Fig.~\ref{fig:vine}. In this architecture, the main chamber and three longitudinal SPM actuators are fabricated together from TPU-coated fabric (30D Nylon 6.6 TPU One sided, Adventure Xpert). The fabric is folded into a double-layer structure and heat sealed using a foot-operated impulse sealer (H-89, ULINE). Longitudinal seals form the body walls, while transverse seals define the individual actuator pouches before the final longitudinal seam is closed to form the 7.62 cm (3 in) diameter tubular body. Using this fabrication process, the resulting vine bodies achieved a burst pressure of 137~kPa (19.9~PSI). 

The integrated construction reduces the amount of fabric required compared with an exterior-pouch design, in which three independently fabricated actuators are attached to the outside of the main chamber. For SPROUT, this reduction in material also decreases the volume of the collapsed body and facilitates packing a longer robot within the available storage volume of the vine base. The integrated-pouch design was selected as a practical balance between growth and steering performance. As characterized in \cite{mcfarland2025onsteerability}, integrated-pouch robots require the pouch pressure to exceed the main chamber pressure before appreciable steering occurs, but they can achieve greater maximum curvature than exterior-pouch robots at higher pressure ratios. The fabrication procedure used for the SPROUT integrated-pouch body is provided in the supplementary documentation.

\section{Demonstration}

Since 2024, SPROUT has been fielded several times at USAR training sites. In particular, we conducted extensive trials of the system at the Massachusetts Task Force 1 training site. This site is an engineered collapsed structure rubble pile with many entry points of varying difficulty to traverse. Under the consultation of technical search specialists, we deployed the system in multiple locations on the pile. Extensive documentation of test performance and evaluation criteria is provided in \cite{mcfarland2024field}, and a recent illustrative set of tests is documented in Fig.~\ref{fig:experiments}. In these trials, the vine base was fitted with a 4 m long body.

The first set of tests (Fig.~\ref{fig:experiments}(a)) focused on the ability of the vine to grow to full length in complex environments. Using 10.3~kPa (1.5~PSI) of chamber pressure, the vine is able to grow at a rate of 0.11~m/s. SPROUT demonstrated reliable growth at inclined angles (Fig.~\ref{fig:experiments}(a)), and into void spaces (Fig.~\ref{fig:experiments}(f)). We tested SPROUT's ability to grow vertically into voids and to steer through a large opening (Fig.~\ref{fig:experiments}(d)) and through a jagged, non-linear entry path (Fig.~\ref{fig:experiments}(c)). Another fundamental skill of vine robots is their ability to grow through apertures smaller than their nominally inflated diameter. In USAR operations, crews drill entry holes 6.4 cm in diameter to allow for manual inspection of voids with search cameras. SPROUT was able to grow up the side of a pile and enter a hole of this diameter in Fig.~\ref{fig:experiments}(e). 

As a further test of this ability, we grew SPROUT through a series of increasingly small holes cut with an exothermic torch into a 0.635 cm thick steel plate (Fig.~\ref{fig:experiments}(b)). We repeated these tests until the vine body could no longer force its way through the opening. The smallest successfully transited hole had a diameter of 3.81 cm -- approximately 50\% smaller than the required diameter. Finally, we deployed the robot through a rubble pile with heterogeneous surface types (Fig.~\ref{fig:experiments}(c)), including organic material, large pieces of broken concrete, steel rebar, and sheet metal.

Over the course of the day during this field campaign, the vine body only experienced a minor puncture after 5 hours of continuous use on abrasive and sharp surfaces. After a puncture or failure like this, the body can be easily replaced onto the vine base to enable continued exploration of the void spaces.

\section{System Extensibility}
The SPROUT platform is not mechanically or computationally restricted to the integrated-pouch  vine body used here. A previous iteration of the system was demonstrated with exterior-pouch bodies in \cite{mcfarland2024field}. Both vine body variants interface with the base through the main body pressure line and three independently controlled pouch motor lines. Beyond these configurations, the compute box can accommodate additional sensing, actuation, or control hardware through its removable electronics plates, available internal space, and existing power and communication interfaces. The platform is also compatible with a tip-based mount \cite{valdivia2026gotta} for the carriage of sensors, or distributed sensing nodes placed along the length of the body \cite{gruebele2021distributed, godoy2026distributed, laudenslager2026evaluatingaccuracyvinerobot}.

\section{Conclusion}
In this paper, we introduced SPROUT, an open-source, open-hardware vine robot designed for field use. Our system introduces a compact, weather-proof compute box for operations in all conditions. We also include a pneumatic and electrical architecture to allow SPROUT to operate independently of hardwired electrical sources or air compressors --- items that are impractical or unrealistic considerations in the field. By building the core software stack around ROS~2, we designed SPROUT to support easy data logging and novel control paradigms. We anticipate that SPROUT will serve as a springboard for other robotics researchers to make advances in soft robotics in the dirty and dangerous conditions where they can effect the most good.


%%%%%%%%%%%%%%%%%%%%%%%%%%%%%%%%%%%%%%%%%%%%%%%%%%%%%%%%%%%%%%%%%%%%%%%%%%%%%%%%

%%%%%%%%%%%%%%%%%%%%%%%%%%%%%%%%%%%%%%%%%%%%%%%%%%%%%%%%%%%%%%%%%%%%%%%%%%%%%%%%

%%%%%%%%%%%%%%%%%%%%%%%%%%%%%%%%%%%%%%%%%%%%%%%%%%%%%%%%%%%%%%%%%%%%%%%%%%%%%%%%

\section*{Acknowledgments}

The authors are thankful to the members of Massachusetts Task Force 1 and Clay Fire Territory for providing access to their training sites and feedback on the robot's capabilities, and to Erin Doran for assistance in improving the vine fabrication.

%%%%%%%%%%%%%%%%%%%%%%%%%%%%%%%%%%%%%%%%%%%%%%%%%%%%%%%%%%%%%%%%%%%%%%%%%%%%%%%%

% \clearpage
% \newpage
\bibliographystyle{IEEEtran} 
\bibliography{references}

@INPROCEEDINGS{Whitman2018uSnake,
  author={Whitman, Julian and Zevallos, Nico and Travers, Matt and Choset, Howie},
  booktitle={IEEE International Symposium on Safety, Security, and Rescue Robotics}, 
  title={Snake Robot Urban Search After the 2017 {M}exico {C}ity Earthquake}, 
  year={2018},
  volume={},
  number={},
  pages={1-6},
  doi={10.1109/SSRR.2018.8468633}}

@article{CoadRAM2020,
  title={Vine Robots: Design, Teleoperation, and Deployment for Navigation and Exploration},
  author = {M. M. Coad and L. H. Blumenschein and S. Cutler and J. A. Reyna Zepeda and N. D. Naclerio and H. El-Hussieny and U. Mehmood and J. Ryu and E. W. Hawkes and A. M. Okamura},
  journal={IEEE Robotics and Automation Magazine},
  year={2020},
    volume={27},
  number={3},
  pages={120--132},
}

@INPROCEEDINGS{Fujikawa2019asc,
  author={Fujikawa, Takumi and Yamauchi, Yu and Ambe, Yuichi and Konyo, Masashi and Tadakuma, Kenjiro and Tadokoro, Satoshi},
  booktitle={IEEE International Symposium on Safety, Security, and Rescue Robotics}, 
  title={Development of Practical Air-floating-type Active Scope Camera and User Evaluations for Urban Search and Rescue}, 
  year={2019},
  volume={},
  number={},
  pages={1-8},
  doi={10.1109/SSRR.2019.8848966}}

@article{Ambe2016Kumamoto,
  title={Use of active scope camera in the {K}umamoto Earthquake to investigate collapsed houses},
  author={Yuichi Ambe and Tomonari Yamamoto and Shotaro Kojima and Eri Takane and Kenjiro Tadakuma and Masashi Konyo and Satoshi Tadokoro},
  journal={IEEE International Symposium on Safety, Security, and Rescue Robotics},
  year={2016},
  pages={21-27},
}

@inproceedings{der2021roboa,
  title={Ro{B}oa: Construction and evaluation of a steerable vine robot for search and rescue applications},
  author={der Maur, Pascal Auf and Djambazi, Betim and Haberth{\"u}r, Yves and H{\"o}rmann, Patricia and K{\"u}bler, Alexander and Lustenberger, Michael and Sigrist, Samuel and Vigen, Oda and F{\"o}rster, Julian and Achermann, Florian and others},
  booktitle={IEEE International Conference on Soft Robotics},
  pages={15--20},
  year={2021}
}

@INPROCEEDINGS{heap2026transparent,
  author={Heap, William E. and Qin, Yimeng and Hammond, Kai and Bayya, Anish and Kong, Haonan and Okamura, Allison M.},
  booktitle={2026 IEEE 9th International Conference on Soft Robotics (RoboSoft)}, 
  title={A Hermetic, Transparent Soft Growing Vine Robot System for Pipe Inspection}, 
  year={2026},
  volume={},
  number={},
  pages={496-503},
  doi={10.1109/RoboSoft67810.2026.11522856}}

@misc{laudenslager2026evaluatingaccuracyvinerobot,
      title={Evaluating Accuracy of Vine Robot Shape Sensing with Distributed Inertial Measurement Units}, 
      author={Alexis E. Laudenslager and Antonio Alvarez Valdivia and Nathaniel Hanson and Margaret McGuinness},
      year={2026},
      eprint={2602.24202},
      archivePrefix={arXiv},
      primaryClass={cs.RO},
      url={https://arxiv.org/abs/2602.24202}, 
}

@inproceedings{el2018development,
  title={Development and evaluation of an intuitive flexible interface for teleoperating soft growing robots},
  author={El-Hussieny, Haitham and Mehmood, Usman and Mehdi, Zain and Jeong, Sang-Goo and Usman, Muhammad and Hawkes, Elliot W and Okamura, Allison M and Ryu, Jee-Hwan},
  booktitle={IEEE/RSJ International Conference on Intelligent Robots and Systems},
  pages={4995--5002},
  year={2018}
}

@article{heap2024large,
  title={Large-Scale Vine Robots for Industrial Inspection: Developing a New Framework to Overcome Limitations With Existing Inspection Methods},
  author={Heap, William E and Man, Steven and Bassari, Vedad and Nguyen, Steven and Yao, Elvy B and Tripathi, Neel A and Naclerio, Nicholas D and Hawkes, Elliot W},
  journal={IEEE Robotics \& Automation Magazine},
  year={2024},
  volume={32},
  number={3},
  pages={64-75},
  publisher={IEEE}
}

@article{valdivia2026gotta,
  title={Gotta Grow Fast: Design and Benchmarking of a Tip Mount for High-Speed Vine Robots},
  author={Alvarez Valdivia, Antonio and Reeve, Robert and Dhawan, Ankush and McFarland, Ciera and Council, Chad and McGuinness, Margaret and Hanson, Nathaniel},
  journal={IEEE Robotics and Automation Letters},
  year={2026},
  publisher={IEEE}
}

@inproceedings{greer2017series,
  title={Series pneumatic artificial muscles ({sPAMs}) and application to a soft continuum robot},
  author={Greer, Joseph D and Morimoto, Tania K and Okamura, Allison M and Hawkes, Elliot W},
  booktitle={IEEE International Conference on Robotics and Automation},
  pages={5503--5510},
  year={2017}
}

@ARTICLE{HawkesScienceRobotics2017,
  author = {E. W. Hawkes and L. H. Blumenschein and J. D. Greer and A. M. Okamura},
  title = {A soft robot that navigates its environment through growth},
  journal = {Science Robotics},
  year = {2017},
  volume = {2},
  pages = {eaan3028},
  number = {8}
}

@inproceedings{gruebele2021distributed,
  title={Distributed sensor networks deployed using soft growing robots},
  author={Gruebele, Alexander M and Zerbe, Andrew C and Coad, Margaret M and Okamura, Allison M and Cutkosky, Mark R},
  booktitle={2021 IEEE 4th International Conference on Soft Robotics (RoboSoft)},
  pages={66--73},
  year={2021},
  organization={IEEE}
}

@article{niiyama2015pouch,
author = {Ryuma Niiyama and Xu Sun and Cynthia Sung and Byoungkwon An and Daniela Rus and Sangbae Kim},
title ={Pouch Motors: Printable Soft Actuators Integrated with Computational Design},
journal = {Soft Robotics},
volume = {2},
number = {2},
pages = {59-70},
year = {2015},
doi = {10.1089/soro.2014.0023},
}

@inproceedings{jeong2020tip,
  title={A tip mount for transporting sensors and tools using soft growing robots},
  author={Jeong, Sang-Goo and Coad, Margaret M and Blumenschein, Laura H and Luo, Ming and Mehmood, Usman and Kim, Ji Hun and Okamura, Allison M and Ryu, Jee-Hwan},
  booktitle={IEEE/RSJ International Conference on Intelligent Robots and Systems},
  pages={8781--8788},
  year={2020}
}

@inproceedings{mcfarland2024field,
  title={Field insights for portable vine robots in urban search and rescue},
  author={McFarland, Ciera and Dhawan, Ankush and Kumari, Riya and Council, Chad and Coad, Margaret and Hanson, Nathaniel},
  booktitle={IEEE International Symposium on Safety Security Rescue Robotics},
  pages={190--197},
  year={2024}
}

@article{mcfarland2025onsteerability,
  title={On Steerability Factors for Growing Vine Robots},
  author={McFarland, Ciera and Alvarez Valdivia, Antonio and Taher, Sarah and Hanson, Nathaniel and McGuinness, Margaret},
  journal={arXiv preprint arXiv:2510.22504},
  year={2025}
}

@book{murphy2017disaster,
  title={Disaster robotics},
  author={Murphy, Robin R},
  year={2017},
  publisher={MIT press}
}

@INPROCEEDINGS{godoy2026distributed,
  title={Distributed Acoustic Localization Array Deployed Using a Soft Everting Vine Robot},
  author={Godoy, Sebastian Lorca and McFarland, Ciera and Val, Michael and Alvarez Valdivia, Antonio and Hanson, Nathaniel and McGuinness, Margaret},
  booktitle = { IEEE/RSJ International Conference on Intelligent Robots and Systems},
  year={2026},
}

@article{macenski2022robot,
  title={Robot operating system 2: Design, architecture, and uses in the wild},
  author={Macenski, Steven and Foote, Tully and Gerkey, Brian and Lalancette, Chris and Woodall, William},
  journal={Science robotics},
  volume={7},
  number={66},
  pages={eabm6074},
  year={2022},
  publisher={American Association for the Advancement of Science}
}

@article{qin20253d,
  title={{3D} Steering and Localization in Pipes and Burrows using an Externally Steered Soft Growing Robot},
  author={Qin, Yimeng and Grinberg, Jared and Heap, William and Okamura, Allison M},
  journal={arXiv preprint arXiv:2507.07225},
  year={2025}
}

% Supplementary Information
\section*{Supplementary Information}

\begingroup
\hypersetup{
    colorlinks=true,
    urlcolor=red,
    linkcolor=red,
    citecolor=red
}
\begin{table}[!h]
    \centering
    % \caption{Description of Supplementary Materials}
    \label{tab:supplementary_materials}

    \renewcommand{\arraystretch}{1.25}

    \begin{tabularx}{\columnwidth}{@{}l X@{}}
        \hline
        \textbf{Material} & \textbf{Description} \\
        \hline

        Documentation &
        Step-by-step instructions showing how to fabricate, assemble, and wire the compute box, vine base, and vine body. Supporting photos and the bill of materials are also provided in the \href{https://sprout-mitll.github.io/sprout/documentation/}{\textcolor{red}{project website}}.\\

        CAD &
        SolidWorks and STEP models for the compute box and robot base. Contains both individual parts for modification and assemblies showing how components fit together.
        Source files are provided in the CAD directory of the \href{https://sprout-mitll.github.io/sprout/cad/}{\textcolor{red}{project website}}. \\

        Electronics &
        Custom Arduino shield PCB design files, including the Gerber files and schematics. Source files are provided in the electronics directory of the \href{https://sprout-mitll.github.io/sprout/electronics/}{\textcolor{red}{project website}}. \\

        Software &
        Instructions and scripts to bootstrap the Jetson computer and Arduino microcontroller with the ROS~2
        software stack. Source files are provided in the software directory \href{https://sprout-mitll.github.io/sprout/software/}{\textcolor{red}{project website}}. \\

        \hline
    \end{tabularx}
\end{table}
\endgroup
\addtolength{\textheight}{-12cm}   % This command serves to balance the column lengths
% \clearpage
% \newpage

\end{document}